\pdfoutput=1

\documentclass[11pt]{article}

\usepackage{acl}

\usepackage{times}
\usepackage{graphicx}
\usepackage{latexsym}
\usepackage{booktabs} 
\usepackage{subfigure}
\usepackage{bbding}
\usepackage{pifont}
\usepackage{multirow}
\usepackage{amsfonts}
\usepackage{color}
\usepackage{amsmath}
\usepackage{microtype}
\usepackage{bbold}
\usepackage{amssymb}
\usepackage[T1]{fontenc}

\usepackage[utf8]{inputenc}

\usepackage{microtype}

\title{Towards Robust Online Dialogue Response Generation}

\author{
  Leyang Cui$^{\heartsuit}$\thanks{\ \ Work was done when Leyang Cui was interning at Pattern Recognition Center, WeChat AI, Tencent.}, \ \ Fandong Meng$^{\spadesuit}$, \ \ Yijin Liu$^{\spadesuit}$, \ \ Jie Zhou$^{\spadesuit}$, \ \ Yue Zhang$^{\heartsuit \Diamond}$ \\
  $^\heartsuit$School of Engineering, Westlake University \\
  $^\spadesuit$Pattern Recognition Center, WeChat AI, Tencent \\
  $^\Diamond$Institute of Advanced Technology, Westlake Institute for Advanced Study \\
 \texttt{\{leyangcui, fandongmeng, yijinliu, withtomzhou\}@tencent.com} \\ \texttt{yue.zhang@wias.org.cn}
  
  }

\begin{document}
\maketitle
\begin{abstract}
Although pre-trained sequence-to-sequence models have achieved great success in dialogue response generation, chatbots still suffer from generating inconsistent responses in real-world practice, especially in multi-turn settings. We argue that this can be caused by a discrepancy between training and real-world testing. At training time, chatbot generates response with the golden context, while it has to generate based on the context consisting of both user utterances and the model predicted utterances during real-world testing. With the growth of the number of utterances, this discrepancy becomes more serious in the multi-turn settings. In this paper, we propose a hierarchical sampling-based method consisting of both utterance-level sampling and semi-utterance-level sampling, to alleviate the discrepancy, which implicitly increases the dialogue coherence. We further adopt reinforcement learning and re-ranking methods to explicitly optimize the dialogue coherence during training and inference, respectively. Empirical experiments show the effectiveness of the proposed methods for improving the robustness of chatbots in real practice. 

\end{abstract}

\section{Introduction}
Sequence-to-sequence neural models \cite{vinyals2015neural} serve as a foundation for dialogue response generation \cite{blender, dialogueGPT},
where typical models adopt the auto-regressive framework \cite{NIPS2014_a14ac55a}.
During training, models are optimized to maximize the token-level likelihood of the golden response given the golden dialogue history context as input; during inference, the dialogue response generation model is required to predict the response token by token based on the golden multi-turn dialogue context. 

\begin{figure}[t!]
    \centering
        \subfigure[Training. \label{fig:1a}]{
    \includegraphics[width=0.5\textwidth]{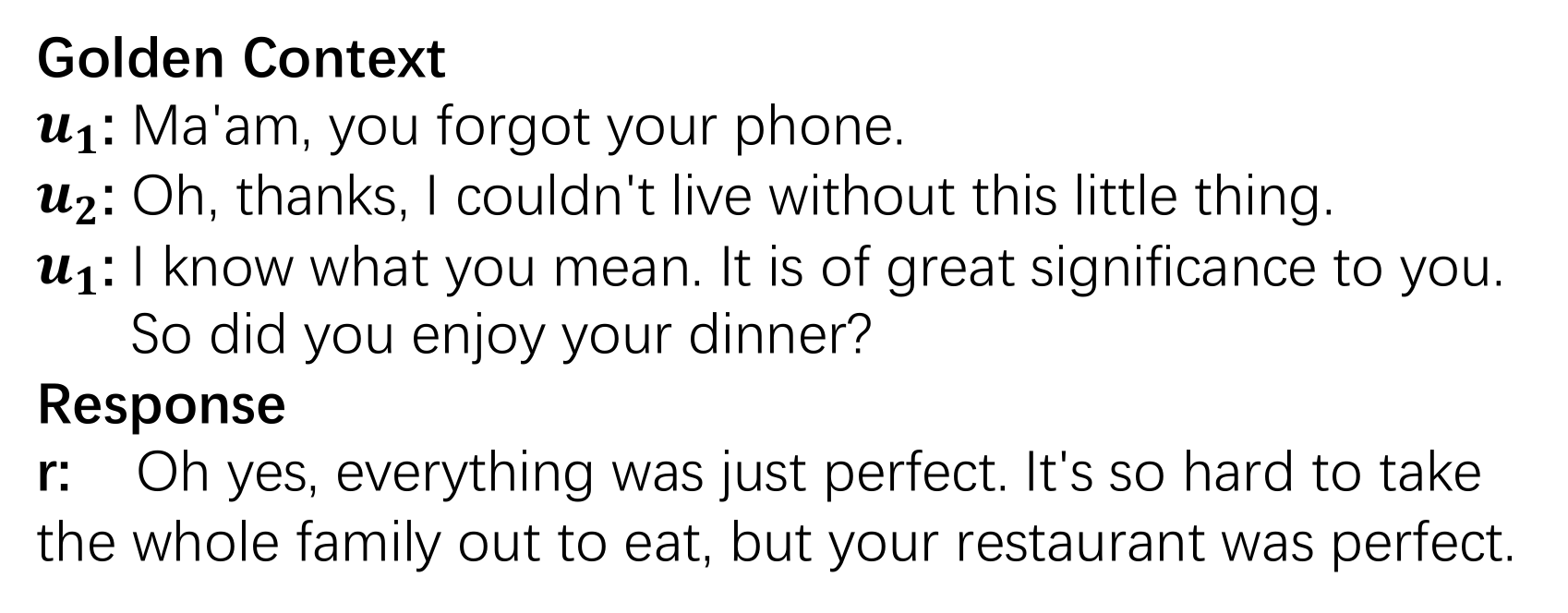}
    }
    \vskip -3pt
        \subfigure[Offline Test. \label{fig:1b}]{
    \includegraphics[width=0.5\textwidth]{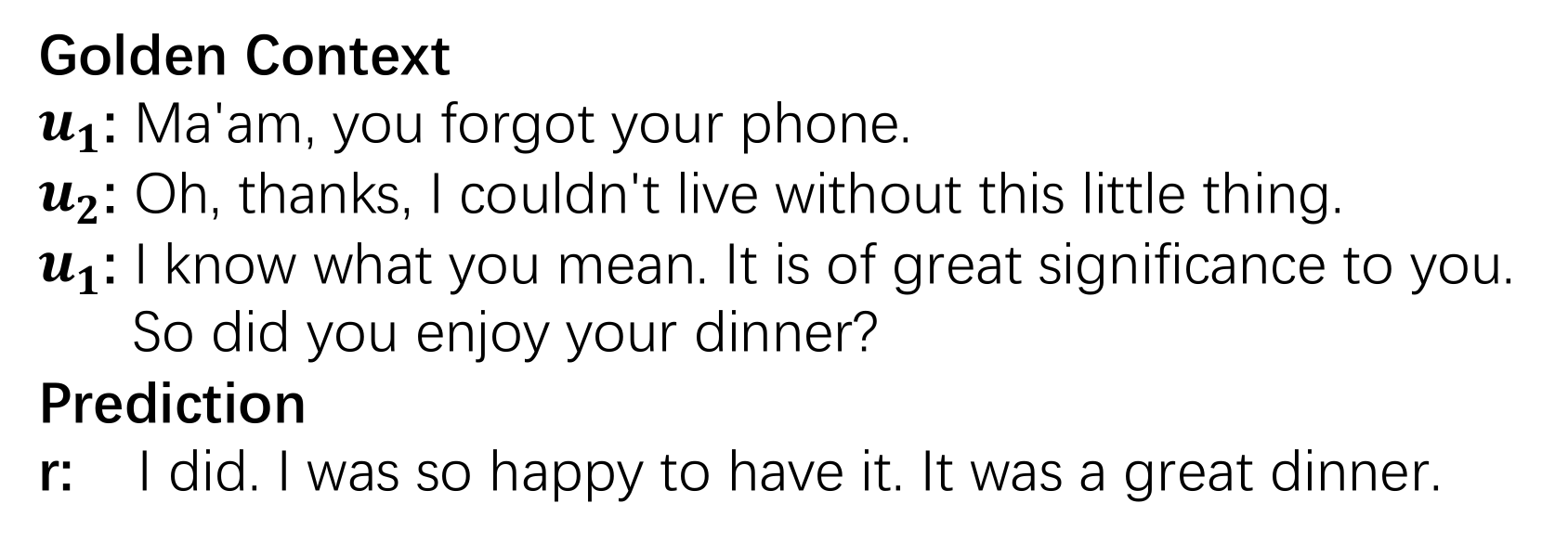}
    }
    \vskip -3pt
    \subfigure[Online Test. \label{fig:1c}]{
    \includegraphics[width=0.5\textwidth]{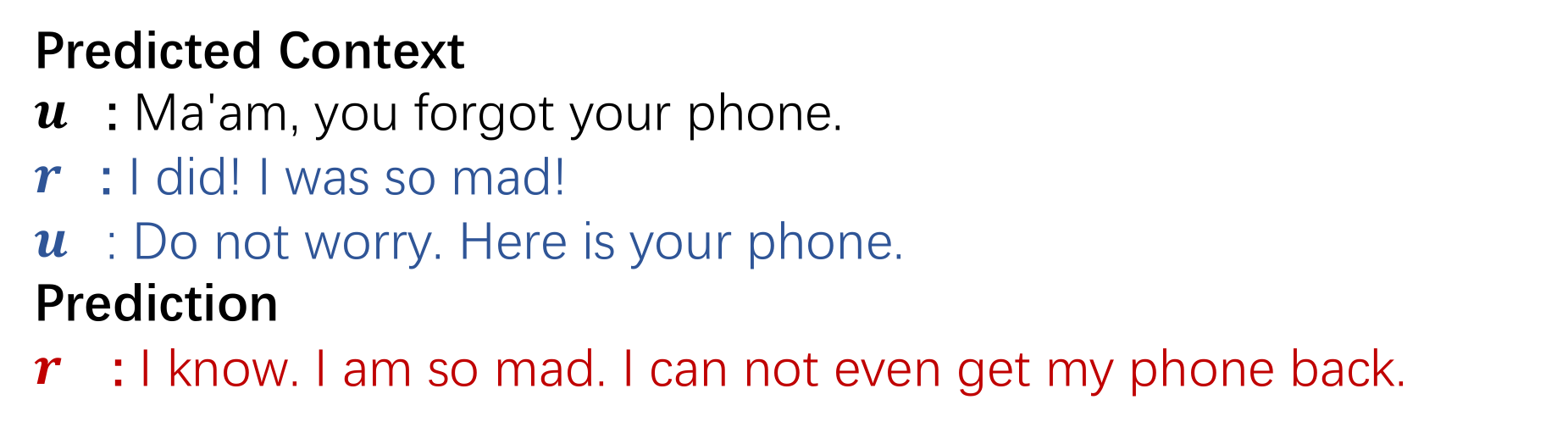}
    }
    \caption{The illustration of how Blender-bot generates responses in different settings. The prompt utterance is sampled from MuTual \cite{mutual}.
    Blender-bot uses golden context in both training and offline test settings.
    The blue part indicates the discrepancy utterances in the context of real-world testing (online test). Blender-bot generates an incoherent response in human-bot conversation (Red utterance in Figure~\ref{fig:1c}).}
    \label{fig:intro}
    \vspace{-0.4cm}
\end{figure}

With advance in large-scale pre-training \cite{dialogpt, blender, bart} and the availability of high-quality conversational datasets \cite{dailydialog, wizard}, models are able to generate fluent and informative responses \cite{xiaoice}. On the other hand, despite achieving promising performance on the standard evaluation metrics (e.g., F-1, BLEU,  PPL), dialogue response generation models still suffer from unsatisfactory user experience in practice \cite{Welleck2020Neural, ram2018conversational}.
Previous work shows that chatbots generate repetition \cite{Unlikelihood} and contradictory responses \cite{decode, aih}.
One possible reason is that current research focuses on the {\it offline} evaluation settings, where the golden context is used as input. However, the golden context cannot be accessed in {\it online} settings.
Figure~\ref{fig:1c} shows a human-bot conversation in practice. 
The golden context in Figure~\ref{fig:1a} and Figure~\ref{fig:1b} is replaced with a system-generated context in Figure~\ref{fig:1c}. In this real-world setting, the multi-turn context consists of both previous chatbot generated utterance ($r$) and human response ($u$), which is inconsistent with the training settings.

\begin{figure}[t!]
    \centering
    \includegraphics[width=0.5\textwidth]{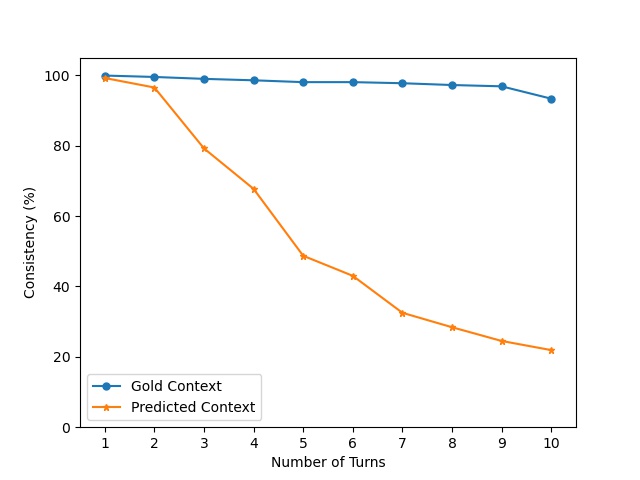}
    \caption{We fine-tune BART on Wizard \cite{wizard} and report the coherence rate against number of utterances on test set. Coherence rate (Eq~\ref{eq:crate}) measures the percentage of responses is coherence with the corresponding contexts.}
    \label{fig:turns}
\end{figure}

Such utterance-level discrepancy between {\it offline} training and {\it online} testing is reminiscent of the exposure bias problem \cite{exposure-bias, exposure-bias-2}. 
Recent research has made solid strides towards alleviating the exposure bias problem in various generation tasks, such as image captioning \cite{exposure-bias}, speech recognition \cite{exposure-bias}, and neural machine translation \cite{bridging, scheduled-transformers}.
They simulate the inference stage by replacing golden target input tokens with the model predictions during training. Intuitively, it can be applied to dialogue generation also.
However, the unique challenge in multi-turn dialogue response generation is the existence of both the utterance-level and token-level discrepancy in a hierarchical manner, which is more severe compared to the above tasks.
Given the golden context, 93.3\% of generated utterances are coherent with the context after 10 turns in our experiments. However, when it comes to the predicted context, the coherence rate drops to less than 30\% (Figure~\ref{fig:turns}).

To alleviate the inconsistency between training and real-world testing, we propose both utterance-level and semi-utterance-level sampling-based methods to improve the performance for the online setting. 
In particular, we sample whole utterances with a scheduled probability and use model generated utterances to replace golden utterances. We schedule our sampling in a hierarchy way.
Utterance-level sampling method generates the utterance based on the previous context, which simulates the online-testing scene during training. 
Semi-utterance-level sampling generates an utterance by using both the previous context and the first few tokens in the sampled utterance, for keeping the semantic similarity between the generated utterance and the golden utterance. 
To further boost the performance, we adopt reinforcement learning and re-ranking to directly optimize the dialogue coherence between the context and the response in the simulated online setting, by consulting an external natural language inference (NLI) based coherence classifier during training and inference, respectively.


We conduct our experiments on Wizard of Wikipedia \cite{wizard} and human-bot conversation.
Empirical results show that our hierarchical sampling approach improves the abilities of dialogue models on generating coherent and less repetitive responses without introducing external training signals. We further demonstrate that an external coherence classifier can be used in both training and inference to help models produce more coherent responses. Finally, we demonstrate that these methods make chatbots more robust in real-word testing.
We release our code and models at \url{https://anonymous}.


\section{Related Work}

\paragraph{Alleviating Discrepancy.}
To bridge the gap between training and inference in auto-regressive models, \citet{exposure-bias} first attempted to randomly sample the previous generated token to replace the ground-truth token during training. \citet{bridging} extended the work of \citet{exposure-bias} by sampling candidates using beam search. \citet{scheduled-transformers} considered scheduled sampling for transformer-based model. \citet{liu-etal-2021-confidence} and \citet{liu_ss_decoding_2021} further designed sampling strategy based on the model confidence and decode steps, respectively. 
\citet{xu2021adaptive} introduced scheduled sampling in the one-to-many generation scenario.
All these method are designed for mitigating the token-level exposure bias problem. To our knowledge, we are the first to improve the utterance-level discrepancy between training and real-world testing.

\paragraph{Dialogue Coherence.}
\citet{dialogue-nli} modeled dialogue coherence as natural language inference and released the dialogue NLI dataset based on persona \cite{persona}. \citet{li-etal-2020-dont} leveraged NLI as supervision to reduce incoherent and repetition response via unlikelihood training.
\citet{decode} extended dialogue NLI by releasing a human-written multi-domain dataset. \citet{qin-etal-2021-dont} further introduced dialogue NLI in task-oriented dialogue system.
\citet{khandelwal-2021-weasul} used reinforcement learning to optimize semantic coherence and consistent flow.
\citet{flat-conversations} proposed a dynamic flow mechanism to model the context flow.
We use coherence as a measure of online dialogue quality. In contrast, existing work all consider the offline setting where the input is a golden history.


\section{Definition}
\subsection{Task}
Given a dialogue context $\mathbf{U} =\{\mathbf{u}_1,\dots,\mathbf{u}_{l-1}\}$, where $\mathbf{u}_i=\{\mathbf{x}^{\mathbf{u}_i}_1,\dots,\mathbf{x}^{\mathbf{u}_i}_{|\mathbf{u}_i|}\}$ represents the $i$-th utterance. $\mathbf{U}$ can be formed as $\mathbf{U}= \{\mathbf{x}_1,\dots,\mathbf{x}_T\}$ by concatenating all utterances as a token sequence, where $\mathbf{x}_i$ denotes the $i$-th token in $\mathbf{U}$.
The corresponding response can be denoted as $\mathbf{r} = \mathbf{u}_{l} = \{y_1, y_2,\dots, y_{T'}\}$.
Given a training context-response pair $\{\mathbf{U}, \mathbf{r}\}$, the probability $P(\mathbf{r}|\mathbf{U})$ can be computed by:
\begin{equation}
\small
    p(\mathbf{r}|\mathbf{U}) = \prod_{t=1}^{T'} p(y_t|\mathbf{U},y_{1:t-1})
\end{equation}
which can be estimated by a sequence-to-sequence neural network (i.e., transformers) with parameters $\theta$.
Our goal is to learn a dialogue generation model $P_\theta(\mathbf{r}|\mathbf{U})$, which is able to generate response $\mathbf{r}$ based on the context $\mathbf{U}$.

\subsection{Model}
\label{sec:model}
We adopt a standard Transformer \cite{transforms} seq2seq model in a dialogue response generation setting.

The dialogue context $\mathbf{U}$ is first fed into the transformer encoder, yielding a sequence of hidden representations.
\begin{equation}
\small
\mathbf{h}^{enc} = \textsc{Transformer\_Encoder}(\mathbf{U}) 
\label{eq:encoder}
\end{equation}

At the $t$ th step of the decoder, $\mathbf{h}^{enc}$ and the previous output tokens $y_{1:t-1}$ are then used as inputs, yielding an output representation
\begin{equation}
\small
\mathbf{h}^{dec}_t = \textsc{Transformer\_Decoder}(\mathbf{h}^{enc}, y_{1:t-1}) 
\end{equation} 

The generative probability distribution of $y_t$ is given by a linear projection of the hidden vector $\mathbf{h}^{dec}_t$ followed by a softmax transformation
\begin{equation}
\small
    p(y_t|\mathbf{U},y_{1:t-1}) = softmax (\mathbf{W}^o \mathbf{h}^{dec}_t + \mathbf{b}^o)
\end{equation}
where $\mathbf{W}^o$ and $\mathbf{b}^o$ are trainable parameters.

The standard cross-entropy loss is used to optimize the parameters $\theta$. Given a training pair $(\mathbf{U},\mathbf{r})$, the objective is to minimize:
\begin{equation}
\small
    \mathcal{L}_{dialogue} = - \sum_{t=1}^{T'} \log p(y_t|\mathbf{U},y_{1:t-1})
\label{eq:dialogue_loss}
\end{equation}

During inference, model auto-regressive generates the response $\mathbf{\hat{r}}$ based on the context $\mathbf{U}$.

\subsection{Evaluation}
\paragraph{Offline Evaluation.}
A conventional practice for evaluating dialogue generation model is formed as a lexical similarity task. In particular, the dialogue generation model is first required to generate response $\hat{\mathbf{r}}$ based on the golden dialogue context $\mathbf{U}$. And then the lexical similarity (i.e., F1, BLEU) between the golden response $\mathbf{r}$ and the generated response $\hat{\mathbf{r}}$ is calculated to measure the performance.


\paragraph{Online Evaluation.}
In real practice, chatbot is used to communicate with human users online. 
As an example for the $l$-th turn, the dialogue context consists of both human utterances and chatbot utterances generated in previous turns, formed as $\mathbf{\hat{U}} =\{\mathbf{u}_1,\hat{\mathbf{r}}_2,\mathbf{u}_3,\hat{\mathbf{r}}_4,\dots,\mathbf{u}_{l-1}\}$, where $\mathbf{u}_i$ represents the $i$-th user utterances and $\hat{\mathbf{r}}_i$ represents the chatbot prediction based on $\mathbf{\hat{U}}_1^{i-1}$. In this setting, the golden context $\mathbf{U}$ does not exist, because the context has been dynamically generated. 
An intuitive method for online evaluation is to employ a human to talk with chatbot naturally. However this evaluation method is high-cost \cite{aih} and relative subjective \cite{convai}, which cannot be adopted in large-scale evaluation.
Following~\citet{spot-the-bot}, we use bot-bot conversations (self-talk) to simulate human-bot conversation, and conduct a NLI-based classifier $f_c(\mathbf{\hat{U}}, \mathbf{\hat{r}})$ to estimate whether the generated response is in line with the context. In particular, given a prompt utterance $\mathbf{u}_1$, we conduct $K$ turns self-talk conversations, yielding a list of utterances $\mathbf{\hat{U}} =\{\mathbf{u}_1,\hat{\mathbf{r}}_2,\hat{\mathbf{r}}_3,\dots,\hat{\mathbf{r}}_K\}$. At turn $k \in [1,K]$, the coherence rate $c_k$ is calculated by:
\begin{equation}
\small
    c_k = \sum_{i=1}^D \frac{\mathbb{1}(f_c(\mathbf{\hat{U}}_1^{i-1}, \mathbf{\hat{r}}_i)=1)}{D}
\label{eq:crate}
\end{equation}
where $D$ represents the number of instances for evaluation, $\mathbb{1}(\cdot)$ returns 1 if $\cdot$ is true and 0 otherwise.


\section{Method}
We take sampling-based methods to simulate online consentaneous (Section~\ref{sec:utterance-sample}), and introduce a reinforcement learning method and a re-ranking method to optimize the dialogue coherence explicitly (Section~\ref{sec:rl-rerank}).

\subsection{Hierarchical Sampling}
The main difference between training and inference in real world practice when generating $\mathbf{\hat{r}}$ is whether we use the golden context $\mathbf{U}$ or the predicted context $\mathbf{\hat{U}}$ partly predicted by the model. We address this by introducing the hierarchical sampling to optimize dialogue coherence implicitly.

\begin{figure}[t!]
    \centering
    \includegraphics[width=0.5\textwidth]{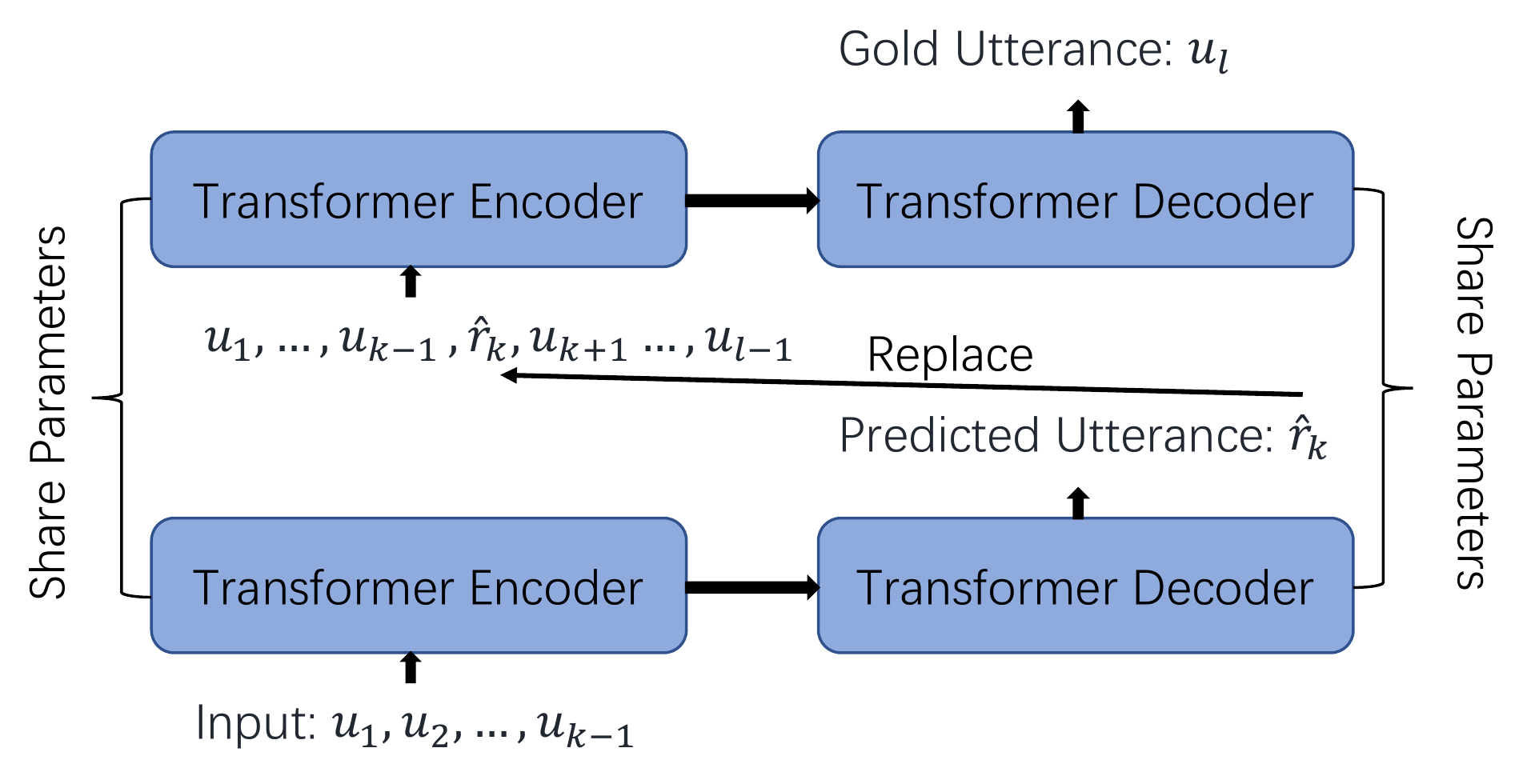}
    \caption{Training with proposed sampling-based methods.}
    \label{fig:sample}
\end{figure}

\label{sec:utterance-sample}
\paragraph{Utterance Level Sampling.}
Our utterance-level sampling mechanism is shown in Figure~\ref{fig:sample}. Given a golden context $\mathbf{U}_1^{l-1}$, we sample an utterance $\mathbf{u}_i, \ i\in [1, l-1]$ from geometric distribution $\sim Geo(p)$ (with $p=0.2$ and max clip $i_{max}=10$), which tends to sample previous utterance to be replaced. After obtaining the utterance $\mathbf{u}_i$, we first ask the model to predict the response $\mathbf{\hat{r}}_i$ based on the previous context $\mathbf{U}_1^{'i-1}$, and then we use the predicted utterance $\mathbf{\hat{r}_i}$ to replace the golden utterance $\mathbf{u}_i$ in the golden context $\mathbf{U}_1^{l-1}=\{\mathbf{u}_1,\dots,\mathbf{u}_i,\dots,\mathbf{u}_{l-1}\}$, yielding the mixed context $\mathbf{U}_1^{'l-1}=\{\mathbf{u}_1,\dots,\mathbf{\hat{r}}_i,\dots,\mathbf{u}_{l-1}\}$. Finally, $\mathbf{U}_1^{'l-1}$ are fed into the encoder. Accordingly, equation~\ref{eq:dialogue_loss} is modified as below:
\begin{equation}
\small
    \mathcal{L}_{dialogue} = - \sum_{t=1}^{T'} \log p(y_t|\mathbf{U}_1^{'l-1},y_{1:t-1})
\label{eq:dialogue_loss_utt}
\end{equation}


\paragraph{Semi-utterance Level Sampling.}

Our semi-utterance-level sampling method generates the response based on both the previous context and the first few tokens in the sampled utterance. In particular, after obtaining the sampled utterance $\mathbf{u}_i$, we further keep the first $j$ tokens in $\mathbf{u}_i$ as additional cues to generate $\mathbf{\hat{r}}'_i$.
Intuitively, a larger $j$ increase both semantic-level and lexical-level overlap between the $\mathbf{\hat{r}}'_i$ and $\mathbf{u}_i$. A smaller $j$ to simulate more accumulate errors along with the inference steps. The same as utterance level sampling in Section~\ref{sec:utterance-sample}, $\mathbf{\hat{r}}'_i$ is used to replace $\mathbf{u}_i$.

\subsection{Explicit Coherence Optimization}
\label{sec:rl-rerank}
\paragraph{Training.}

Inspired by optimizing the generation towards high BLEU\cite{yang-etal-2018-improving}, we introduce a reinforcement learning method, which explicitly optimizes the coherence between the context and the generated response. We fine-tune the dialogue model $P_\theta$ to optimize the reward model $P^{RL}_\theta$.

\begin{figure}[t!]
    \centering
        \subfigure[Reward Calculation. \label{fig:rla}]{
    \includegraphics[width=0.3\textwidth]{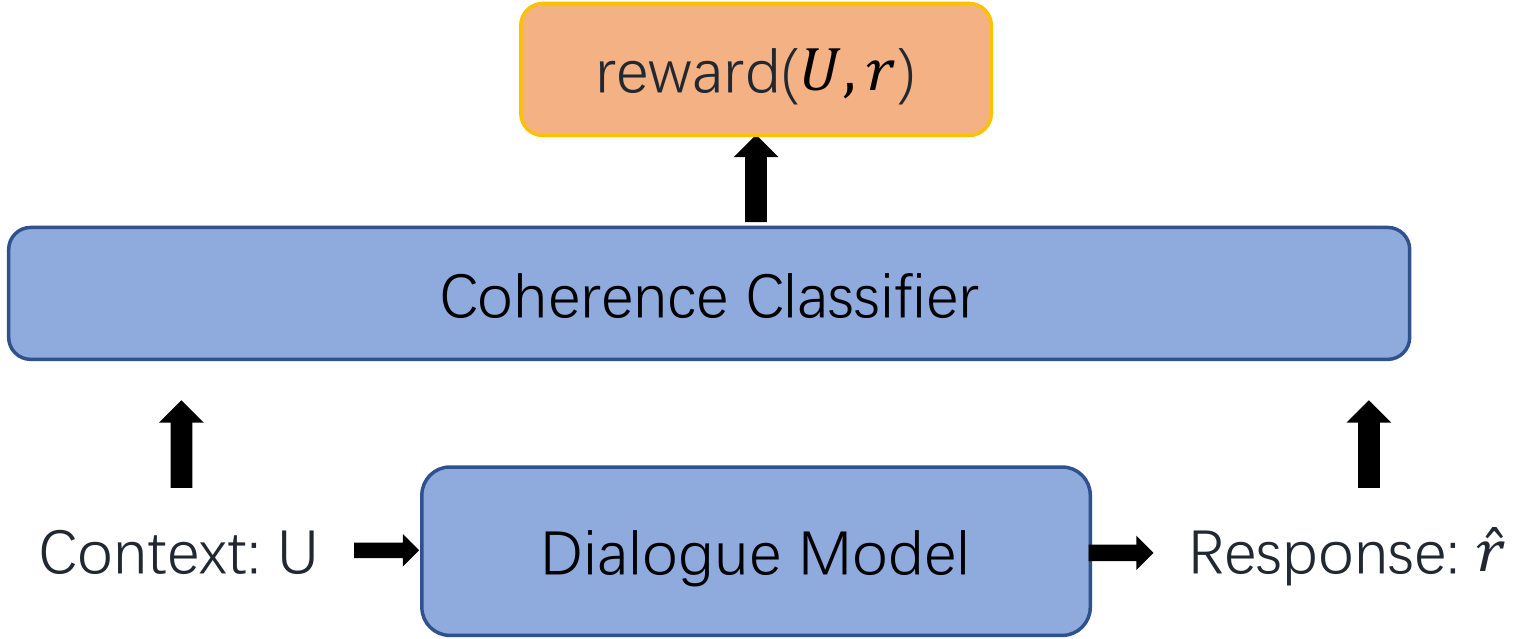}
    }
        \subfigure[Optimization. \label{fig:rlb}]{
    \includegraphics[width=0.5\textwidth]{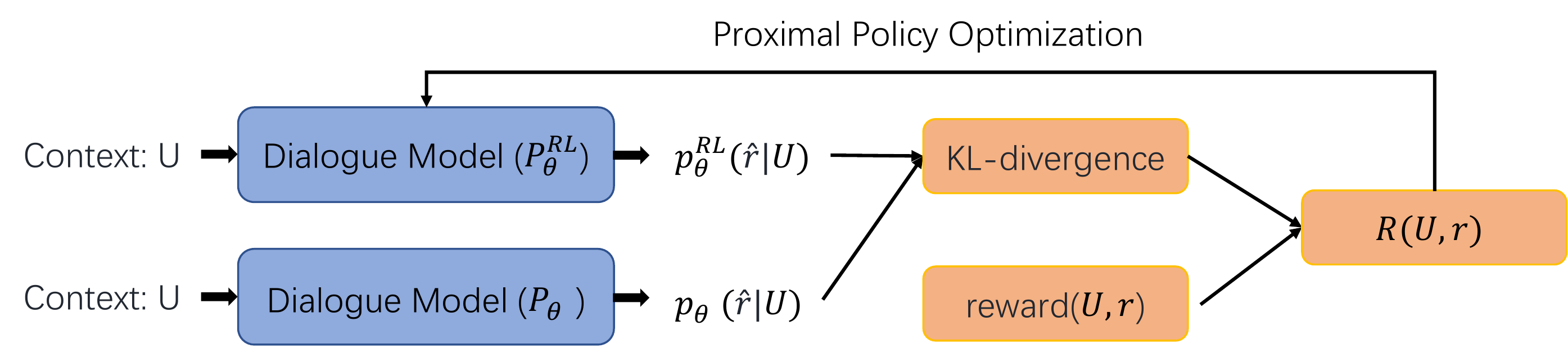}
    }
    \caption{Coherence-Oriented Reinforcement Learning.}
    \label{fig:rl}
\end{figure}

As shown in Figure~\ref{fig:rla}, we first ask the model to generate a response $\mathbf{\hat{r}}$ based on the context $\mathbf{U}$. 
Then an external coherence classifier $f_c$ is used to justify whether the response is coherent with the context. We adopt the logits of $f_c$ corresponding to the coherent label as the reward. In particular, the input of $f_c$ is a context-response pair $(\mathbf{U}, \mathbf{r})$ and the output is whether the response is coherent with the context.
For training $f_c$, we turn context-response pair $(\mathbf{U}, \mathbf{r})$ to {\tt [CLS]} $\mathbf{U}$ {\tt [SEP]} $\mathbf{r}$ {\tt [SEP]}, and feed it into the RoBERTa model. The hidden state of the {\tt [CLS]} token is used for MLP followed by a softmax scoring function to obtain the coherence score.
We train $f_c$ on {\bf D}ialogu{\bf E} {\bf CO}ntradiction {\bf DE}tection (DECODE) \cite{decode}, which is a human annotated corpus labeled with ``contradiction (non-coherent)'' and ``non-contradiction (coherent)''. The classifier achieves 94.24 on DECODE dev.

Following~\citet{ziegler2019finetuning} and \citet{jaques2020way}, we additionally introduce a Kullback–Leibler (KL) divergence term to prevent $P^{RL}_\theta$ from drifting too far from $P_\theta$ (Figure~\ref{fig:rlb}). Formally, given the context $\mathbf{U}$, we calculate the KL-divergence between two models' output probabilities 
\begin{equation}
    KL(\mathbf{U}) = \sum_{t=1}^{T'} \log \frac{p^{RL}_\theta(\mathbf{x}_t|\mathbf{U},\mathbf{x}_{1:t-1})}{p_\theta(\mathbf{x}_t|\mathbf{U},\mathbf{x}_{1:t-1})}
\end{equation}

$KL(\mathbf{U})$ can be considered as a KL-divergence for the language model task.

Finally, we optimize $P^{RL}_\theta$ using Proximal Policy Optimization (PPO) \cite{ppo} with the clipped reward:
\begin{equation}
    Reward(\mathbf{U, \mathbf{r}}) = f_c(\mathbf{U}, \mathbf{\hat{r}}) - \beta KL(\mathbf{U})
\label{eq:rl-reward}
\end{equation}
where $\beta$ is a hyper-parameter to control the contribution of the KL term.
Intuitively, we use the classifier to encourage the model to generate coherent responses, and rely on the KL term to ensure fluency. The inference stage can be the same as the baseline methods in Section~\ref{sec:model}.

\paragraph{Inference with Re-ranking.}
Another method to enhance dialogue coherence explicitly is inference with re-ranking.
In particular, we first adopt beam search to produce multiple candidates responses, and then re-rank the utterances using the coherence classifier $f_c$. At each turn, the candidate with the highest coherence score is used as the response.

\section{Experiments}

We train our model based on the golden context - response pair on Wizard of Wikipedia \cite{wizard}, a chit-chat dialogue benchmark. 
Two annotators are employed to chat based on an initial topic. 
The dataset contains 18,430 training dialogues with 1,365 topics.

\subsection{Metrics}
Following \citet{wizard} and \citet{skt}, the perplexity (PPL) of the ground-truth response, given the golden context as input is taken as one automatic metric. Additionally, coherence rate and non-repetition rate are used as automatic metrics, and human evaluation is conducted.

\paragraph{Coherence Rate.}
To evaluate online performance in real-world practice, we conduct self-talk to simulate the human-bot conversation, and measure whether the generated response is coherent with the previous context as one automatic metric. The maximum interaction turn is set to 10.
As model-based methods have been proved efficient and reliable \cite{decode,ke-blender,aih}, and we evaluate the dialogue coherence by consulting $f_c$ in Section~\ref{sec:rl-rerank}.

\paragraph{Non-Repetition Rate.}
Inspired by \citet{diversity}, we adopt non-repetition rate to quantify the diversity of the generated sequence during self-talk as a second automatic metric. We calculate distinct-1, distinct-2 and distinct-3 by counting the diversity of uni-grams, bi-grams and tri-grams, respectively. For each context $\mathbf{\hat{U}}$, the distinct-$n$ is calculated by:
\begin{equation}
    \text{distinct}-n = \frac{\textsc{count}(\textsc{unique}_{\text{$n$-gram}_i \in \mathbf{\hat{U}}}(\text{$n$-gram}_i))}{\textsc{count}(\textsc{total}_{\text{$n$-gram}_i \in \mathbf{\hat{U}}}(\text{$n$-gram}_))}
\end{equation}
where \textsc{count}(), \textsc{unique}() and \textsc{total}() denote count the item of a list, unique items in a list and enumeration a list, respectively.
A higher distinct-$n$ indicates a lower repetition rate during self-talk.

\begin{table*}[t!]
    \centering
    \small
    {\resizebox{0.96\textwidth}{!}{
    \begin{tabular}{c|cccccccccccc|c}
    \toprule
    & \multicolumn{12}{c|}{{Online Evaluation}} & Offline \\
    \midrule
         & $c_1$ & $c_2$ & $c_3$ & $c_4$ & $c_5$ & $c_6$ & $c_7$ & $c_8$ & $c_9$ & $c_{10}$ & avg\_5 & avg\_10 & PPL \\
    \midrule
         BART w/ Golden context & 99.7 & 98.9 & 98.2 & 96.0 & 97.6 & 97.2 & 96.0 & 94.2 & 94.1 & 93.3 & 99.0 & 96.5 & - \\
    \midrule
         Single-turn BART & 99.2 & 88.1 & 71.5 & 63.5 & 57.2 & 53.0 & 46.7 & 41.8 & 37.3 & 34.9 & 75.9 & 59.3 & 21.3 \\ 
         Multi-turn BART & 99.2 & 96.5 & 79.2 & 67.7 & 48.7 & 43.0 & 32.5 & 28.4 & 24.5 & 21.9 & 78.3 & 54.2 & 17.8 \\
    \midrule
         w/ Noise & 99.2 & 95.4 & 76.5 & 58.7 & 47.1 & 35.4 & 31.4 & 22.1 & 23.1 & 12.4 & 75.4 & 50.1 & 18.1 \\
         w/ Utterance & 98.4 & 97.0 & 89.3 & 76.7 & 71.6 & 59.1 & {\bf 60.5} & {\bf 45.7} & {\bf 49.8} & {\bf 35.6} & 86.6 & {\bf 68.4} & 17.2 \\
         w/ Semi-Utterance & 98.1 & 97.2 & 85.7 & 69.2 & 64.0 & 50.5 & 52.1 & 36.4 & 43.6 & 29.1 & 82.9 & 62.6 & {\bf 17.1} \\
         w/ Hierarchical & {\bf 99.2} & {\bf 97.6} & {\bf 91.2} & {\bf 78.5} & {\bf 72.3} & {\bf 60.7} & 57.8 & 45.5 & 44.3 & 33.0 & {\bf 87.8} & 68.0 & 17.4  \\
    \bottomrule
    \end{tabular}
    }}
    \caption{Test performance of self-talk given a prompt utterance on Wizard test set.}
    \label{tab:main-results}
    \vspace{-7pt}
\end{table*}

\paragraph{Human Evaluation.}
Following previous work~\cite{human-eval}, we conduct human evaluation on self-talk to compare our hierarchical sampling-based methods with our baseline multi-turn BART by randomly sampling 50 instances (including 500 utterances).
Following~\citet{wu2018response}, we employ three annotators to do a side-by-side human evaluation.

In order to pursue more authentic evaluation in real practice, we further adopt a human-bot conversation to online evaluate these two methods. In particular, given a prompt utterance, we ask an annotator to chat with chatbot 10 turns. The final human-bot test set we derive contains 50 dialogues (including 500 utterances) for each model.
We define three metrics for human evaluation, including fluency, non-repetitive and coherence. Each aspect is scored into three grades (0, 1 and 2) representing ``bad'', ``normal'' and ``good'', respectively. We further calculate the Pearson correlation between the human annotated coherence rate and the model assigned coherence rate.



\subsection{Baselines}
We compare the proposed methods with the following BART-based baselines:
\paragraph{BART w/ Golden context.}
We fine-tune BART on the Wizard training set. During inference at turn $k$, the golden context $\mathbf{U}_1^{k-1}$ is used to produce the response $\mathbf{\hat{r}}_k$. Because the golden context is unavailable in practice, the performance can be considered as the ceiling performance for alleviating the discrepancy between training and real-world testing.

\paragraph{Multi-turn BART.}
During training, we fine-tune BART based on the golden context-response pair. Different from BART w/ Golden context, we use the context $\mathbf{\hat{U}}_1^{k-1}$ predicted by previous turns to generate the response $\mathbf{\hat{r}}_k$ during inference.

\paragraph{Single-turn BART.}
We fine-tune BART for the dialogue generation following the single-turn setting \cite{single-turn}. Only the last predicted utterance $\hat{\mathbf{r}}_{k-1}$ is fed to the encoder to generate $\mathbf{\hat{r}_k}$ for both training and inference. Single-turn BART ignores the history in previous utterances.

\paragraph{w/ Noise}
After sample an utterance $\mathbf{u}_i$, we use a random noise $\mathbf{u}_{random}$ randomly sampled from the training set to replace $\mathbf{u}_i$.

\subsection{Results}
Table~\ref{tab:main-results} reports the performance of coherence rate as well as PPL for various methods, and Table~\ref{tab:distinct} shows the distinct-$n$ for the predicted context generated by these methods.

\paragraph{Predicted Context vs Golden Context.}
We first compare whether the dialogue generation model is able to generate coherence response based on the golden context and the predicted context.
As shown on the top of Table~\ref{tab:main-results}, the coherence rate of BART w/ Golden context does not decrease significantly with the number of turns increasing. The performance drops by only 5.6 points coherence rate from 2 turns to 10 turns. However, given the predicted context, the coherence rate decreases sharply as the number of turns increase, with only 21.9 $c_{10}$. This shows the severity of the discrepancy problem in real-world multi-turn dialogue generation.

\paragraph{Single-turn vs Multi-turn.}
In {\it offline} evaluation, multi-turn BART achieves 17.8 PPL, which significantly outperforms single-turn BART. This indicates that context information is important for response generation. However, we have mixed results in {\it online} evaluation. For example, multi-turn BART outperforms single-turn BART when the number of utterances in the context is less than four in Table~\ref{tab:main-results}. 
When the number of utterances becomes larger, single-turn BART surprisingly gives better results compared with multi-turn BART. The reason can be that the mismatch between the golden context and the predicted context hinders the model performance as the number of utterances grows for multi-turn model. 

\paragraph{Sampling vs w/o Sampling.}
In Table~\ref{tab:main-results}, the proposed sampling-based approach performs slightly better on PPL compared to the multi-turn BART, which shows our methods also work well in general offline settings. When it comes to online settings, our sampling-based methods outperform multi-turn BART significantly in all metrics, although there is no direct supervision signal on coherence. For example, when measured in context corresponding to 5 turns, multi-turn BART w/ hierarchical sampling gives a $c_5$ of 72.3\%, as compared to 48.7\% by multi-turn BART. Furthermore, multi-turn BART w/ Noise do not work well, since sampled noises are difficult to accurately simulate errors of the inference scene during training.


\begin{table}[t!]
    \centering
    \small
    \begin{tabular}{c|c|c|c}
    \toprule
    Model & Dis-1 & Dis-2 & Dis-3 \\
    \midrule
    Multi-turn BART & 24.37 & 32.30 & 36.35 \\
    w/ Hierarchical sampling & 36.29 & 49.77 & 55.29  \\
    \midrule
    \end{tabular}
    \vspace{-0.1cm}
    \caption{Non-Repetition Rate (\%) for $n$-gram. `Dis-$n$' means `Distinct-$n$'.}
    \label{tab:distinct}
\end{table}


\begin{table}[t!]
    \centering
    \small
    \begin{tabular}{c|c|c|c}
    \toprule
    
         Model & Fluency & Rep & Coh  \\
    \midrule
        \multicolumn{4}{c}{{Self-talk}} \\
    \midrule
         Multi-turn BART & {\bf 1.93} & 0.89 & 0.74 \\
         w/ Hierarchical sampling & 1.91 & {\bf 1.37} & {\bf 1.45}  \\
    \midrule
            \multicolumn{4}{c}{{Human-bot Conversation}} \\
    \midrule
        Multi-turn BART & 1.89 & 0.96 & 0.63 \\
         w/ Hierarchical sampling & {\bf 1.90} & {\bf 1.53} & {\bf 1.32} \\
    \bottomrule
    \end{tabular}
    \caption{Human Evaluation. `Rep' and `Coh' indicate non-repetition and coherence, respectively.}
    \label{tab:eval-human}
\end{table}

\begin{figure}[t!]
    \begin{center}
    \subfigure[Multi-turn BART. \label{fig:rerank1}]{
    \includegraphics[width=0.45\textwidth]{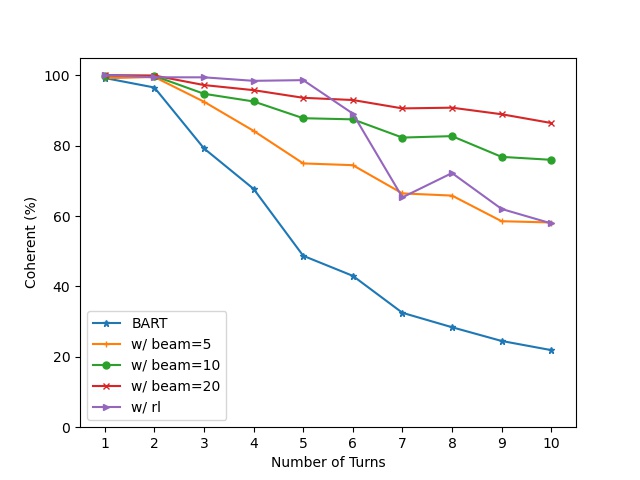}
    }
    \subfigure[Multi-turn BART w/ Hierarchical Sampling. \label{fig:rerank2}]{
    \includegraphics[width=0.45\textwidth]{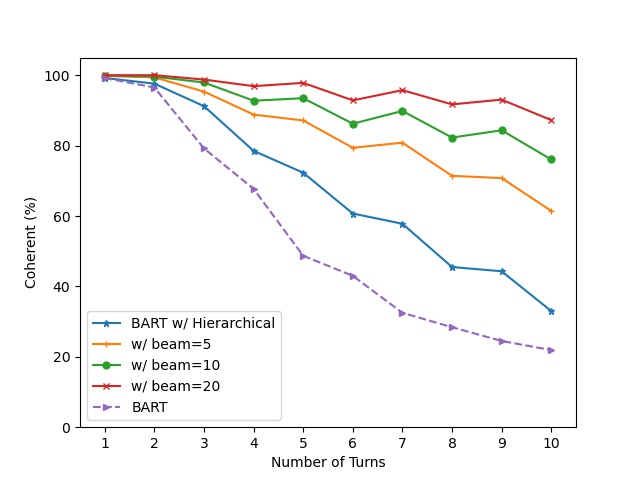}
    } 
    \caption{Coherence rate with explicit optimization.}
    \label{fig:rerank}
    \end{center}
\end{figure}

\paragraph{Utterance vs Hierarchical.}
In Table~\ref{tab:main-results}, semi-utterance level sampling underperforms utterance-level sampling in online evaluation. This is because semi-utterance level sampling cannot accurately simulate errors of the inference scene during training. For instance, the dialogue model tends to generate the response beginning with the word ``${\textit I}$''. While semi-utterance level sampling keeps the first few tokens in the sampled utterance. When integrating utterance-level and semi-utterance level sampling, hierarchical sampling gives the best coherence rate when context less than six turns, which achieves 87.8\% on $avg_5$. This shows the effectiveness of sampling in a hierarchy way, which simulates the errors on both utterance-level and token-level.

\paragraph{Repetition.}
Table~\ref{tab:distinct} reports the non-repetition rate of our sampling-based methods, drawing multi-turn BART as a reference. We find that our methods give higher distinct-$n$ measured by uni-gram, bi-gram and tri-gram, which shows the effect of introducing hierarchical sampling to reduce copying and repetition in model generated context. This also provides support for the effectiveness of sampling-based methods to increase the robustness of multi-turn models.

\paragraph{Human Evaluation.}
Table~\ref{tab:eval-human} compares the hierarchical sampling-based method with multi-turn BART using human evaluation. All models are able to produce fluent responses due to the power of pre-training, where fluency exceeds 1.89 for all models. Measured in non-repetition and coherence, our hierarchical sampling method significantly outperforms the baselines ($p$\textless0.01) on both self-talk and human-bot conversation. In human-bot conversation, the coherence increases largely from 0.96 to 1.53, showing that sampling enhances the robustness of online multi-turn conversation.
For self-talk, the pearson correlation between the human annotated and the model assigned coherence rate is 0.78, which also demonstrates the effectiveness of the model-based evaluation methods.


\begin{table*}[t!]
    \centering
    \small
    \begin{tabular}{c|l}
    \toprule
    $\mathbf{u}_1$(Prompt) & My favorite video game is Quake. Have you ever played it? \\
    \midrule
    \multicolumn{2}{c}{\bf Multi-turn BART} \\
    \midrule
    $\mathbf{\hat{r}}_2$ & I have not played it, but I know it was developed by the Quake team. \\
    $\mathbf{\hat{r}}_3$ & Oh, Quake was developed by the Quake team. It's a great game!	\\
    $\mathbf{\hat{r}}_4$ & I know it was developed by the Quake team. It was the first video game to be released commercially. \\
    \midrule
        \multicolumn{2}{c}{\bf Multi-turn BART w/ Hierarchical sampling} \\
    \midrule
    $\mathbf{\hat{r}}_2$ & I have not played it, but I have heard it is a very good game. \\	
    $\mathbf{\hat{r}}_3$ & Yes it is. It was developed by the Quake team. It's a team-based game. \\
    $\mathbf{\hat{r}}_4$ & That sounds like a fun game to play with friends. What other games do you like? \\
    \bottomrule
    \end{tabular}
    \vspace{-0.2cm}
    \caption{Examples of generated responses given a prompt utterance on the Wizard of Wikipedia Test Seen.}
    \label{tab:case1}
    \vspace{-0.2cm}
\end{table*}

\paragraph{Explicit Objectives.}
Figure~\ref{fig:rerank} shows the effect of the explicit coherence optimization method. 
Training model with reinforcement learning outperforms with MLE measured by coherence rate, showing the usefulness of optimizing the dialogue coherence directly.
We also find that the coherence rate improves significantly after re-ranking in the inference scene for both multi-turn BART and multi-turn BART w/ hierarchical sampling. 
Furthermore, as the number of candidate utterances increases, the coherence rate increases. Multi-turn BART w/ beam=20 even achieves 86.42 $c_{10}$ compared with 21.9 $c_{10}$ for multi-turn BART.
This indicates that the dialogue model can give coherent response candidates, which can be re-ranked by an external coherence classifier to produce a coherent response. Our hierarchical sampling-based methods also consistently perform better than multi-turn BART by introducing coherence re-ranking. 

\begin{figure}[t!]
    \centering
    \includegraphics[width=0.4\textwidth]{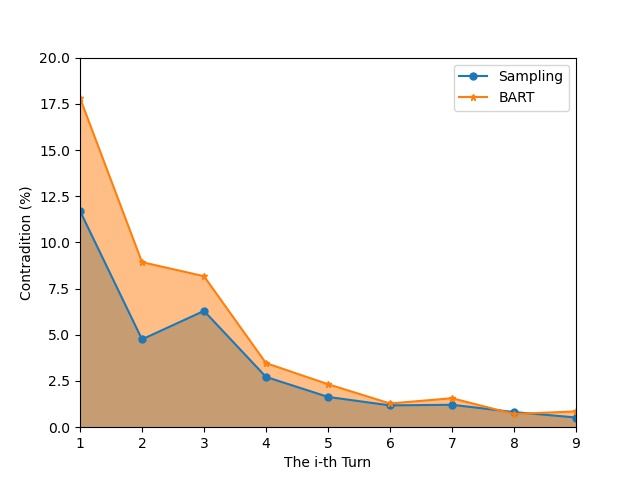}
    \caption{Contradiction rate across different turn. Contradiction rate defined by (1 $-$ coherence rate) $\times 100$\%.}
    \label{fig:contra}
\end{figure}

\begin{figure}[t!]
    \centering
    \includegraphics[width=0.45\textwidth]{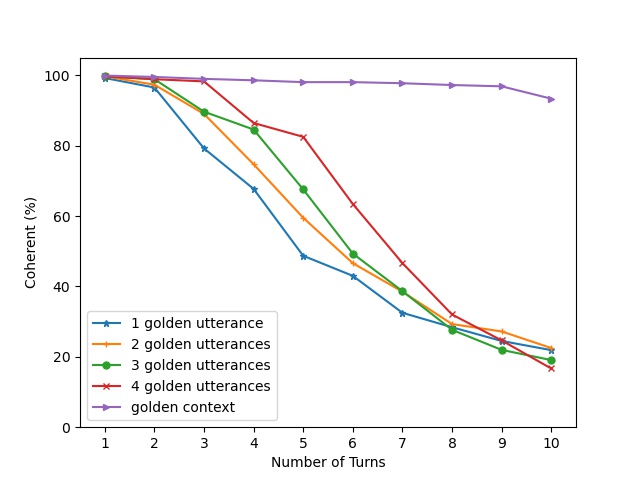}
    \caption{Coherence rate across the number of golden utterances at the beginning.}
    \label{fig:goldturn}
\end{figure}

\section{Analysis}
\paragraph{The Number of Golden Turns.}
We investigate whether a larger number of golden turns at the start is able to help model to produce more coherent responses during inference.
Figure~\ref{fig:goldturn} shows the coherence rate against the number of golden utterances at the beginning during the self-talk, drawing using the golden context as a reference.  
It can be seen that a larger number of golden utterance at the beginning yields a larger coherence rate in the first few turns. However, the coherence rate decreases sharply with the number of turns increasing, which shows that simply increasing beginning golden turns cannot help to alleviate the discrepancy between training and real-world testing. 

\paragraph{Utterance-level Contradiction.}
To understand which turns in the context leads to an incoherence response, we introduce an utterance-based classifier to probe different utterances during generating the response at $10$-th turn in self-talk. As shown in Figure~\ref{fig:contra}, both models tend to generate response that contradict with the early turns. This shows that current models do not take full advantage of the long-range dialogue context. Compared with the multi-turn BART, the proposed sampling-based methods significantly decrease the contradiction rate in the early turns, and achieves the similar results in the later turns, which shows our hierarchical sampling-based methods are able to improve robustness of multi-turn models by alleviating the error accumulation.
 
\paragraph{Case Study.}
We present an example to better understanding of multi-turn BART and our model in Table~\ref{tab:case1}. We observe that both models are able to generate reasonable response $\hat{\mathbf{r}_2}$. Because the context for generating $\hat{\mathbf{r}_2}$ contains prompt utterance (golden context) $\mathbf{u}_1$ only. However, when the model encounters the predicted utterance as context, multi-turn BART tends to generate response with repetition and contradiction. With hierarchical sampling, our model produces coherence responses during self-talk. 

\section{Conclusion}
We quantified online dialogue generation in practice, and proposed the hierarchical sampling-based methods to alleviate the discrepancy between training and real-world testing. 
We further introduce an external coherence classifier on both training and inference to boost the performance. 
Experiments demonstrate the effectiveness of our methods for generating robust online response on both self-talk and human-bot conversation. 

\section{Acknowledge}
We would like to thank Zhen Yang and Sen Yang their helpful discussion and valuable feedback.

\bibliography{anthology, custom}
\bibliographystyle{acl_natbib}

\appendix

\section{Appendix}
\label{sec:appendix}

\subsection{Setup}
We implement our methods with {\tt transformers} and choose {\tt bart-base} as the pre-trained transformer language model. AdamW \cite{adamw} with a batch size of 32 is used to optimize parameters. The initial learning is set as 5e-5, which will be halved in each training iteration. Following~\citet{bart}, we set the maximum input tokens as 512.
The training time of our methods is 0.6 times slower than the baseline method.
Our inference time is the same as that of the baseline.
For the coherence-oriented reinforcement learning method, we set $\beta$ in Equation~\ref{eq:rl-reward} as 0.2. For computational efficiency, we truncate the maximum decode length as 20 to calculate the KL-divergence.

\end{document}